\documentclass[preprints,article,accept,pdftex,oneauthor]{Definitions/mdpi}

\firstpage{1}
\pubvolume{1}
\issuenum{1}
\articlenumber{0}
\pubyear{2026}
\copyrightyear{2026}
\datereceived{ }
\daterevised{ } %
\dateaccepted{ }
\datepublished{ }

\usepackage{subcaption}

\Title{Analysis of Numerical Localisation in LLM Translations}   %
\Author{Patrizia Kaye $^{1}$\orcidA{}}                           %
\AuthorNames{Patrizia Kaye}                                      %
\corres{Correspondence: patrizia.kaye@dunelm.org.uk}             %
\address[1]{$^{1}$ \quad University of Bath; pjk49@bath.ac.uk}   %

\abstract{
	The work of \citeauthor{Tang2025} on numerical translation is extended by analysing the capability of five large language models (LLMs) for the \textit{localisation} of times, numbers, and dates instead of translation.
	Models were selected that could be loaded onto and run on commodity hardware and a baseline quality for each mode is computed, then three different strategies to improve on that accuracy were tested.
	In contrast to \citeauthor{Tang2025}, it was discovered that on the tested LLMs, embedding the localisation principles into the prompt context provided a statistically significant improvement in accuracy compared to direct translation or the alternative strategies.
}


\begin{document}
	\section{Introduction}

Natural languages are fundamentally different to programming languages: while the latter are precisely, formally defined, the former varies by the speaker, their location, when they speak, and so on.
In natural languages words can change their meaning over time and mean different things between cultures and countries, for example: ``chips'' in the UK has the same meaning as ``fries'' in the US, with the US ``chips'' meaning ``crisps'' in the UK.
Consequently, any linguistic model of a language will be an approximation at best, and perfect results cannot be expected.
Even if a model was developed that gave perfect accuracy at a given moment in time, that accuracy would decay, since languages evolve.

Since the initial release of ChatGPT by OpenAI in November 2022, interest in Large Language Models (LLMs) -- as well as AI in general -- has grown rapidly, turning them into an extremely popular source of discussion, both in the general public and amongst the academic community.
One area in which LLMs excel is in translation of one human language to another -- they are now the most popular method for text translation, and are widely favoured compared to machine-translation efforts\cite{Kocmi2024, Kocmi2025}.

However, translation is more than just word selection: it also encompasses ``localisation'': a set of conventions for formatting of information like dates, times, and numbers that varies by country, language, and region.
Using the UK and USA for another example; confusion between the UK's DD/MM/YYYY vs. the USA's MM/DD/YYYY date formats clearly shows why accuracy in this area is important.
In this paper, we consider translation and localisation between English and German; as shown in Table \ref{tab:format_examples}, German uses a period (.) as a thousands separator and a comma as the decimal marker, whereas in the UK this is reversed.
If the number \texttt{123.456} were improperly localised when translating from English to German, whatever quantity was involved would be incorrect by three orders of magnitude!
For any kind of economic or scientific measurement, this would make the translation highly inaccurate, or even nonsensical.
English and German were chosen as the language pair due to the wide availability of high-quality parallel corpora, and the author's knowledge of both languages, which eased development and validation of results.

An under-studied area is how this translation ability applies to numbers, with the only direct study being \citeauthor{Tang2025}'s, which analysed English $\longleftrightarrow$ Chinese translation of numbers in a variety of contexts\cite{Tang2025}.
This paper tests the strategies described by \citeauthor{Tang2025} (see section \ref{sec:strategies}) as they apply to numerical \textit{localisation} instead of translation.
Although a niche area, it is another facet that should be considered when translating text, as expectations for formatting and localisation of values vary between countries, cultures, and languages.

\subsection{Contributions}
	\begin{enumerate}
		\item Analysis of English $\longleftrightarrow$ German numerical localisation using two generic, and three translation-specific LLMs.
		\item Determination of the most accurate method of ensuring localisation accuracy.
		\item A reproducible methodology that is applicable to additional language pairs.
	\end{enumerate}

\subsection{Scope}
	Models were selected for evaluation that can run on commodity hardware (8GB of VRAM and 64GB system memory), reflecting what many organisations and users have available.
    These hardware constraints imposed a model size limit of $\sim$8B parameters and a practical limit of 100 samples for each data type.

    The results do not necessarily generalise to models beyond those that were tested.
    This is a consequence of models being updated with different methods of training or being released with different architectures, however, the scripts and code associated with this framework are available, making directly-comparable analysis of additional and future models feasible.

	\section{Literature Review}

\subsection{Large Language Models}

    A foundational paper was \citeauthor{Vaswani2017}'s proposal of the ``Transformer'' architecture\cite{Vaswani2017}.
    Prior to this, most LLMs had a recurrent neural network (RNN) component alongside an attentional mechanism, but the authors made a convincing case that the RNN component is not in fact required.
    
    In recurrent and convolutional neural networks, the number of calculations required to model a dependency between two tokens increases as the distance between them increases.
    The Transformer architecture reduces this to a constant number of computations irrespective of the distance between the tokens by encoding their \textit{relative} positions, then using a multi-head attention mechanism where each head processes different projected values of the tokens.
    State-of-the-art results were achieved on the WMT14 BLEU metric for English to German and English to French translation with only 10\% of the training costs of previous models.

    In the seminal paper \textit{Improving Language Understanding by Generative Pre-Training}\cite{Radford2018} models were pre-trained using unsupervised learning on a vast corpus of unlabelled text to provide a good initial base for supervised fine-tuning for specific tasks.
    This allowed them to create a generic base model that could be used as the starting point for more specialised training for specific tasks.
    Their results showed that the use of texts with long-range dependencies (i.e. texts longer than phrases and sentences, such as paragraphs and pages), combined with a Transformer architecture provide the best performance.

\subsection{Machine Translation}

    Rule-based machine translation (RBMT) is often used when working with standardised documents (e.g. technical manuals) which require format consistency and technical precision.
    However, RBMT is less useful for unstructured text; \citeauthor{Tao2025} focused on the example of battlefield reports, which can often contain nuance and contextual ambiguities.
    They created a fine-tuning translation dataset for generic LLMs consisting of term-, sentence-, and paragraph-level translations from a high-quality, expert-translated bilingual corpora\cite{Tao2025}.
    Several popular Llama, Qwen, and Deepseek LLMs with $\le$ 10b parameters (to keep within the limits of commodity hardware) were tested and found to show significant improvements to RBMT across each of the tested metrics and converge within a similar time-frame.
    Despite their results, the authors noted deploying such a system in production would require robust systems to mitigate the risk of hallucinations in classified document translation and to comply with relevant data governance standards.

    The BLEU metric has a critical shortcoming when it comes to the use of LLMs; it does not recognise equivalences between words or syntactic constructions, such as using parentheses instead of a pair of commas to indicate an aside.
    As a result, \textit{BLEURT} has been proposed as an end-to-end learned metric, to replace the hand-crafted BLEU metric\cite{Sellam2020}.
    The rationale was that BLEU does not closely correspond to human assessment, whereas learned metrics have greater scope for interpolation over gaps in training data and can be tuned according to the needs of the application to favour faithfulness or a particular style.
    Synthetic data was generated for a pre-training phase by perturbing Wikipedia articles, dropping words, and round-trip translating and the models were further trained on task-specific (translation or data-to-text) data.
    Consistent improvements were achieved in the WMT17-20 benchmarks and the pre-training was found to increase reliability with low-quality data.

\subsection{Numerical Translation}

    Despite its ubiquity in real-world applications, numerical translation remains only lightly researched.
    The most direct work on the matter is \citeauthor{Tang2025}'s \citeyear{Tang2025} paper, appropriately titled \textit{Investigating Numerical Translation with LLMs}.
    The authors describe the state of translation of integers, decimals, numerals and separators between Chinese and English, explaining that no model is completely accurate, whether commercial or open-source, and that while most models exhibited an average accuracy in the low-90\% range, the lowest recorded accuracy was with the Bloomz-7b-mt model at 83\% and 76\% for en $\rightarrow$ cn and cn $\rightarrow$ en translations respectively.
    
    A dataset was created consisting of ten types of numerical translation, from large units, ranges, and numerical strings (e.g. employee numbers), to ratios, formulae, and negative numbers.
    Across the range of tested models none had uniformly better performance, even when this would be intuitive; for instance, the \texttt{Llama3.1} model with 8b parameters was better at translating ratios and formulae than the variant with 70b parameters, although the latter performed much better with fractions and decimals.
    
    Three different approaches to improve the reliability of numerical translations were evaluated:
    \begin{itemize}
    	\item In-context learning (ICL): including principles of conversion within the prompt.
    	\item Chain-of-thought (CoT): prompting the model to explain its process (noting that this is generated text, and need not correspond to the steps actually taken for the translation).
    	\item Post-editing (PE): extracting numerical pairs and prompting separately for their translation.
    \end{itemize}
    ICL typically only corrected a single error, and CoT would often have incorrect translations used during the explained process, but that were correct in the final stage, whereas PE consistently outperformed both ICL and CoT.
    The authors noted that the problems with ICL and CoT highlight the inherent limits of an LLM to understand numerical calculations, whereas PE focuses on direct translation/correction of pre-existing text, a task for which LLMs are highly effective.

    Whilst this appears to be the only work focusing on a systematic study of numerical translation, the localisation of values to the target language was outside the scope of the work.
    It is this gap in the literature that this paper aims to help fill.

\subsection{Datasets}

    Machine translation datasets vary from monolingual to parallel and have a wide variety of registers, ranging from informal, monolingual comments drawn from social media, to formal speeches given by government ministers.
 	The following datasets cover different levels of formality and structure whilst still being available as parallel corpora, which allows for validation of LLM-produced values with known-good examples.
	\texttt{Statmt-europarl-10-deu-eng}\cite{dataset-Europarl-2020}: text extracted from the proceedings of the European Parliament.
    It is formal and covers a wide range of topics.
	\texttt{Tilde-ema-2016-deu-eng}\cite{dataset-EMA-2017}: text retrieved from the European Medicines Agency document portal.
    The text is formal and technical.
	\texttt{Statmt-news\_commentary-18.1-deu-eng}\cite{dataset-NewsCommentary}: an intermediate level of formality, covering a wide variety of topics.

	\section{Method}

Experiments were performed to determine whether LLMs were able to \textit{localise} numbers to the target language as effectively as they translated text.
All experiments were run on a system equipped with an Nvidia GeForce 4060 with 8GB VRAM and 64GB of system memory.
Models were loaded and run using version 0.17.0 of the \texttt{vllm} Python library\cite{Kwon2023}.
Prompts were iteratively adjusted during initial development to improve output, clarify formatting, and remove extraneous text: systematic optimisation would require significant model-specific tuning.

\subsection{Datasets}
	The following datasets were selected for use and downloaded using the \texttt{MTData} tool\cite{gowda-etal-2021-many}.
	They cover different levels of formality and structure whilst still being available as parallel corpora, which allows for validation of the LLM-produced values with known-good examples.
	\begin{itemize}
		\item \texttt{Statmt-europarl-10-deu-eng}\cite{dataset-Europarl-2020} -- 100 samples were taken of each of time, date, and numbers.
		\item \texttt{Tilde-ema-2016-deu-eng}\cite{dataset-EMA-2017} -- 100 samples were taken of each of dates and numbers. The presence of large numbers of titration ratios caused too many false positives when attempting to extract examples of times.
		\item \texttt{Statmt-news\_commentary-18.1-deu-eng}\cite{dataset-NewsCommentary} -- 100 samples were taken of each of dates and numbers. Fewer than 100 instances of times were available in this dataset given the means of extraction, so for consistency between results these were not analysed.
	\end{itemize}
    In total, there were 700 inputs for each model.
    100 samples was deemed the minimum viable number of samples for each data type/dataset combination for the model results to become clear.

\subsection{Data Formats}
	Dates (both numerical and textual), numbers, and times were used as the formats for localisation analysis.
	Some examples of how they differ between English and German are shown in Table \ref{tab:format_examples}.
	In German, formatting typically follows DIN-5008\cite{DIN-5008-web}, which is aligned to ISO-8601-1\cite{ISO-8601-1:2019} but with additional allowances, such as allowing the \texttt{DD.MM.YYYY} date format.
	In English, ISO-8601-1 is also acceptable for formal and data interchange formats, but conventional use has more flexibility\cite{CLDR}.
	\begin{table}[h]\centering
		\caption{Illustrative examples of data formats.}
		\begin{tabular}{c | c | c}
			\textbf{Type} & \textbf{English} & \textbf{German}  \\
			\hline
			Date          & 3$^{rd}$ July 1986       &  3. Juli 1986            \\
			              & 3/7/1986                 &  3.7.1986                \\
			              & 3/7/1986                 &  1986-7-3                \\

			Number        & 123,456                  &  123.456                 \\
			              & 5.6789                   &  5,6789                  \\

			Time          & 5:30 PM                  &  17:30 Uhr               \\
			              & 9:15 AM                  &  9:15 Uhr                \\
			              & 9 AM                     &  9 Uhr                   \\
		\end{tabular}
		\label{tab:format_examples}
	\end{table}

\subsection{Models}
	The following models were selected as they are freely-available and can all run on commodity hardware.
	Two generic models were selected alongside the translation-specific ones, to see if specific training improved the outcome.
	All models are instruction-tuned and use the ChatML template structure for prompts.
	\begin{enumerate}
		\item Generic: \texttt{Qwen2-7B-Instruct}\cite{qwen2}
		\item Generic: \texttt{Qwen3-4B-Instruct-2507}\cite{qwen3technicalreport}
		\item Translation-specific: \texttt{TowerInstruct-7B-v0.2}\cite{tower_llm_2024}
		\item Translation-specific: \texttt{SalamandraTA-7b-instruct}\cite{Salamandra2025}
		\item Translation-specific: \texttt{SalamandraTA-2b-instruct}\cite{Salamandra2025}
	\end{enumerate}
	Models were configured with a temperature of $0$ -- since each sample was processed once, it was important to measure the model's most-probably token output, rather than take a single random sample from its probability distribution.

\subsection{Strategies}
\label{sec:strategies}
	First, direct translation was performed to set a baseline quality against which the improvements (if any) offered by the other strategies could be measured.
	The models were simply instructed to translate the source sentence into the target language.
	Three strategies were assessed in an attempt to find a better way to ensure values are localised correctly when translated.
	\begin{itemize}
		\item \textit{In-context learning}: the model is provided a number of hints to guide the localisation, such as the digit and decimal separators in each of the languages.

		\item \textit{Chain-of-thought}: the model instructed to explain the process behind the translation.
		As noted in \cite{Tang2025}, the generated text does not necessarily reflect the steps actually taken by the model.

		\item \textit{Post-editing}: regular expressions were used to extract numeric pairs from the results of direct translations. The extracted pairs were then translated via the \texttt{dateparser} python module and reinserted into the sentence to provide the final post-edited translation. Where the direct translation failed to produce values, these were excluded from the post-editing results so as to avoid bias.
	\end{itemize}

\subsection{Pipeline}
\label{sec:pipeline}
	First, each dataset was read and each line checked using regular expressions to see if it contained a date, time, or large number (one that had thousands separators or decimal values).
    Incorrectly-formatting data would therefore be excluded by implication, allowing the experiments to focus on known-good data without risking results being confounded by edge-cases and noise.
	Lines with none of these were ignored, but any line matching a particular regular expression would be added (alongside the corresponding line in the other language) to be added to that dataset's collection, for example, the EMA's list of dates or the European Parliament's list of times.
	This resulted in a list of corresponding sentences containing a localisable data type from which 100 random samples were taken (the seed was not recorded, but the samples are available in the supplementary data, allowing reproduction from the same input data).
	This process ensured identical input across all models, allowing for consistent comparison.

	Second, each model was prompted for each dataset, data type, and direction (English $\longleftrightarrow$ German) and the raw responses stored.
	For post-editing, the direct translation was used as the base text.

	Stage three was the cleanup and collation of data: the raw prompt responses were preserved for transparency, and then underwent some simple automated processing to remove artifacts that escaped suppression by the prompt template, for example, preceding the output by ``Here is the translation:''.
	For each prompt and response, the original and translated sentences were grouped together, along with a list of all the data pairs, ready for review and analysis.

    Each pair was automatically compared against the authoritative translation and was awarded up to two point:
    one for the correct value, and one for the correct localisation format.
    Date and time values were checked by the \texttt{dateparser} python library, which allows for matching unknown formats: due to the locale ambiguity, both US and UK formats were accepted as valid.
    Numerical value parsing was handled by the \texttt{babel} library.
	All data formats were checked by \texttt{babel}, which allowed for validating a format against either the \texttt{en} or \texttt{de} locale, as appropriate.
    In line with the scope of this study, whether text was deemed to be localised correctly depending solely on the value pairs, even if the surrounding translation contained inaccuracies.

    The pipeline was then repeated, but explicitly specifying the locale, i.e. prompting for translations into "English (en\_GB)" and "German (de\_DE)".
	See Appendices for the regex patterns prompt templates.
	The full source test setup, \texttt{vllm} configuration and sampled sentences can be found at \url{https://doi.org/10.5281/zenodo.21797172}.

	\section{Results}

The \texttt{Qwen3-4B-Instruct-2507} model outperformed all others, achieving the highest accuracy for each data type and direction irrespective of the strategy used.
Table~\ref{tab:accuracies} shows that the best-performing strategy was in-context learning, which had the highest mean accuracy whether the locale was unspecified or explicit.

A Wilcoxon test was used to determine that the difference in accuracy between in-context learning and other strategies were statistically significant for both unspecified and explicit locales ($\alpha=8.33\times10^{-3}$).
The p-values shown in Tables~\ref{tab:wilcoxon_unspecified} and~\ref{tab:wilcoxon_explicit}.
\begin{table}[H]\centering
    \normalsize
    \caption{Mean accuracy per strategy across all tests.}
    \begin{tabular}{c c c}
        \textbf{Strategy}      & \textbf{Unspecified locale} & \textbf{Explicit locale} \\
        \hline
        Direct                & 50.3\%       & 49.8\%     \\
        In-context learning   & 53.3\%       & 53.8\%     \\
        Chain of thought      & 40.2\%       & 40.3\%     \\
        Post-editing          & 26.9\%       & 28.4\%     \\
    \end{tabular}
    \label{tab:accuracies}
\end{table}

\begin{table}[H]\centering
    \normalsize
    \caption{Wilcoxon p-values for unspecified locales, six pairwise comparisons, $\alpha = 8.33\times10^{-3}$.}
    \begin{tabular}{c | c c c}
        Strategy & ICL                   & CoT                   & PE \\
        \hline
        Direct   & 1.06e-02              & 7.67e-02              & 1.60e-05 \\ 
        ICL      & \cellcolor{lightgray} & 1.37e-04              & 1.02e-07 \\ 
        CoT      & \cellcolor{lightgray} & \cellcolor{lightgray} & 8.14e-03 \\ 
    \end{tabular}
    \label{tab:wilcoxon_unspecified}
\end{table}

\begin{table}[H]\centering
    \normalsize
    \caption{Wilcoxon p-values for explicit locales, six pairwise comparisons, $\alpha = 8.33\times10^{-3}$.}
    \begin{tabular}{c | c c c}
        Strategy & ICL                   & CoT                   & PE \\
        \hline
        Direct   & 4.03e-03              & 7.67e-02              & 8.33e-07 \\ 
        ICL      & \cellcolor{lightgray} & 2.77e-03              & 9.31e-09 \\ 
        CoT      & \cellcolor{lightgray} & \cellcolor{lightgray} & 2.99e-03 \\ 
    \end{tabular}
    \label{tab:wilcoxon_explicit}
\end{table}

Full results can be seen in tables~\ref{tab:results_unspecified} and~\ref{tab:results_explicit}, where the following formatting is used to highlight value of interest:
\begin{itemize}
    \item \colorbox{blue!10}{Cell shading}: indicates a model's most successful strategy for localising a particular data type and direction.
    \item \textbf{Bold typeface}: indicates the best performing model/strategy combination across all combinations for a particular data type and direction.
\end{itemize}

\subsection{Unspecified Locale}
	\begin{table}[H]\centering
    \footnotesize
    \caption{Per-model accuracy of English $\longleftrightarrow$ German localisation across all datasets.}
    \begin{tabular}{l | c  c | c  c | c  c}
                    & \multicolumn{2}{c|}{times}  & \multicolumn{2}{c|}{dates}  & \multicolumn{2}{c}{numbers}   \\
        model / strategy       & en $\rightarrow$ de    & de $\rightarrow$ en                                   & en $\rightarrow$ de    & de $\rightarrow$ en                                   & en $\rightarrow$ de    & de $\rightarrow$ en \\
    \hline
        \texttt{Qwen2-7B-Instruct / direct} & 70.2 & 64.8 & 73.5 & 53.4 & 80.3 & 83.3  \\
        \texttt{Qwen2-7B-Instruct / icl} & \cellcolor{blue!10} 70.7 & 69.9 & \cellcolor{blue!10} 76.9 & \cellcolor{blue!10} 60.1 & \cellcolor{blue!10} 82.8 & \cellcolor{blue!10} 85.3  \\
        \texttt{Qwen2-7B-Instruct / cot} & 54.0 & \cellcolor{blue!10} 72.5 & 29.0 & 24.4 & 52.6 & 53.3  \\
        \texttt{Qwen2-7B-Instruct / pe} & 62.2 & 69.7 & 20.4 & 42.2 & 48.6 & 54.5  \\
    \hline
        \texttt{Qwen3-4B-Instruct-2507 / direct} & 72.9 & \cellcolor{blue!10} \textbf{74.7} & \cellcolor{blue!10} \textbf{81.4} & 64.1 & \cellcolor{blue!10} \textbf{92.2} & \cellcolor{blue!10} \textbf{87.9}  \\
        \texttt{Qwen3-4B-Instruct-2507 / icl} & \cellcolor{blue!10} \textbf{75.2} & 71.8 & 79.4 & \cellcolor{blue!10} \textbf{67.9} & 91.7 & 87.5  \\
        \texttt{Qwen3-4B-Instruct-2507 / cot} & 53.8 & 62.8 & 17.7 & 16.4 & 52.1 & 52.0  \\
        \texttt{Qwen3-4B-Instruct-2507 / pe} & 73.4 & 74.5 & 18.6 & 49.5 & 49.4 & 50.0  \\
    \hline
        \texttt{TowerInstruct-7B-v0.2 / direct} & 20.5 & 46.1 & 58.5 & 26.7 & \cellcolor{blue!10} 74.0 & \cellcolor{blue!10} 66.5  \\
        \texttt{TowerInstruct-7B-v0.2 / icl} & \cellcolor{blue!10} 51.3 & \cellcolor{blue!10} 60.9 & \cellcolor{blue!10} 60.5 & 30.8 & 67.8 & 61.8  \\
        \texttt{TowerInstruct-7B-v0.2 / cot} & 32.5 & 47.4 & 55.8 & \cellcolor{blue!10} 41.9 & 72.7 & 64.7  \\
        \texttt{TowerInstruct-7B-v0.2 / pe} & 44.2 & 17.4 & 11.9 & 20.4 & 29.2 & 32.2  \\
    \hline
        \texttt{salamandraTA-2b-instruct / direct} & \cellcolor{blue!10} 30.2 & 19.0 & 41.9 & 24.5 & 22.0 & 36.5  \\
        \texttt{salamandraTA-2b-instruct / icl} & 25.2 & 19.9 & 42.3 & 25.5 & 31.0 & \cellcolor{blue!10} 42.6  \\
        \texttt{salamandraTA-2b-instruct / cot} & 27.6 & \cellcolor{blue!10} 23.9 & \cellcolor{blue!10} 45.5 & \cellcolor{blue!10} 33.2 & \cellcolor{blue!10} 34.2 & 40.2  \\
        \texttt{salamandraTA-2b-instruct / pe} & 4.2 & 15.5 & 1.5 & 2.3 & 1.1 & 1.5  \\
    \hline
        \texttt{salamandraTA-7b-instruct / direct} & 35.7 & 12.9 & 17.5 & 4.1 & 43.1 & 30.4  \\
        \texttt{salamandraTA-7b-instruct / icl} & \cellcolor{blue!10} 38.0 & \cellcolor{blue!10} 18.7 & \cellcolor{blue!10} 20.6 & 6.1 & \cellcolor{blue!10} 46.8 & \cellcolor{blue!10} 31.2  \\
        \texttt{salamandraTA-7b-instruct / cot} & 34.7 & 13.0 & 15.7 & \cellcolor{blue!10} 8.4 & 43.3 & 29.3  \\
        \texttt{salamandraTA-7b-instruct / pe} & 6.2 & 4.7 & 0.0 & 0.3 & 0.6 & 1.8  \\
    \end{tabular}
    \label{tab:results_unspecified}
\end{table}

	\begin{figure}[H]\centering
		\includegraphics[width=\textwidth]{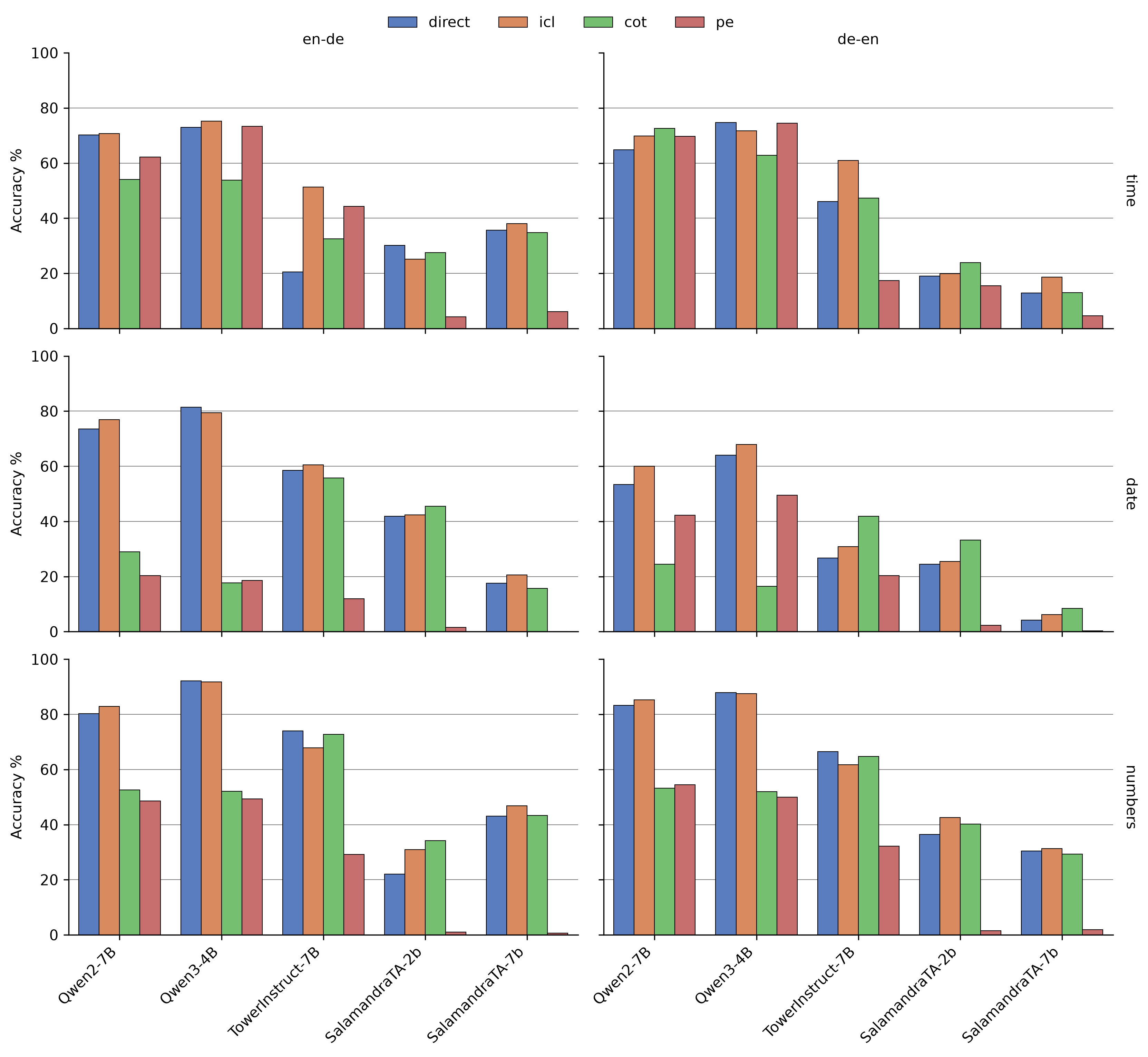}
		\caption{Accuracy for each model and strategy with unspecified locale.}
		\label{fig:results_unspecified}
	\end{figure}

\subsection{Explicit Locale}
	\begin{table}[H]\centering
    \footnotesize
    \caption{Per-model accuracy of English (en\_GB) $\longleftrightarrow$ German (de\_DE) localisation across all datasets.}
    \begin{tabular}{l | c  c | c  c | c  c}
                    & \multicolumn{2}{c|}{times}  & \multicolumn{2}{c|}{dates}  & \multicolumn{2}{c}{numbers}   \\
        model / strategy       & en $\rightarrow$ de    & de $\rightarrow$ en                                   & en $\rightarrow$ de    & de $\rightarrow$ en                                   & en $\rightarrow$ de    & de $\rightarrow$ en \\
    \hline
        \texttt{Qwen2-7B-Instruct / direct} & 71.5 & 64.1 & 73.0 & 55.8 & 82.6 & 83.0  \\
        \texttt{Qwen2-7B-Instruct / icl} & \cellcolor{blue!10} 72.6 & 72.1 & \cellcolor{blue!10} 75.5 & \cellcolor{blue!10} 60.1 & \cellcolor{blue!10} 85.2 & \cellcolor{blue!10} 85.3  \\
        \texttt{Qwen2-7B-Instruct / cot} & 49.6 & 73.6 & 30.5 & 25.6 & 52.5 & 52.7  \\
        \texttt{Qwen2-7B-Instruct / pe} & 61.9 & \cellcolor{blue!10} 74.5 & 33.5 & 42.6 & 49.0 & 57.6  \\
    \hline
        \texttt{Qwen3-4B-Instruct-2507 / direct} & 74.2 & 37.7 & \cellcolor{blue!10} \textbf{81.4} & \cellcolor{blue!10} \textbf{74.9} & 92.0 & \cellcolor{blue!10} \textbf{88.7}  \\
        \texttt{Qwen3-4B-Instruct-2507 / icl} & \cellcolor{blue!10} \textbf{76.8} & 64.0 & 78.0 & 73.3 & \cellcolor{blue!10} \textbf{92.6} & 79.1  \\
        \texttt{Qwen3-4B-Instruct-2507 / cot} & 54.3 & 61.3 & 22.2 & 19.3 & 53.5 & 51.8  \\
        \texttt{Qwen3-4B-Instruct-2507 / pe} & 39.4 & \cellcolor{blue!10} \textbf{75.6} & 38.5 & 49.4 & 51.5 & 49.9  \\
    \hline
        \texttt{TowerInstruct-7B-v0.2 / direct} & 23.7 & 43.0 & \cellcolor{blue!10} 60.8 & 27.4 & \cellcolor{blue!10} 72.7 & \cellcolor{blue!10} 66.3  \\
        \texttt{TowerInstruct-7B-v0.2 / icl} & \cellcolor{blue!10} 49.0 & 45.4 & 59.8 & 30.9 & 70.2 & 64.0  \\
        \texttt{TowerInstruct-7B-v0.2 / cot} & 25.6 & \cellcolor{blue!10} 46.1 & 58.0 & \cellcolor{blue!10} 45.2 & 69.7 & 62.6  \\
        \texttt{TowerInstruct-7B-v0.2 / pe} & 39.9 & 43.8 & 14.1 & 22.3 & 34.7 & 31.1  \\
    \hline
        \texttt{salamandraTA-2b-instruct / direct} & \cellcolor{blue!10} 43.2 & 22.3 & 30.5 & 26.9 & 18.4 & 39.5  \\
        \texttt{salamandraTA-2b-instruct / icl} & 30.6 & 21.8 & \cellcolor{blue!10} 53.0 & 28.1 & \cellcolor{blue!10} 40.8 & \cellcolor{blue!10} 43.7  \\
        \texttt{salamandraTA-2b-instruct / cot} & 30.4 & \cellcolor{blue!10} 23.8 & 48.3 & \cellcolor{blue!10} 36.7 & 36.9 & 41.1  \\
        \texttt{salamandraTA-2b-instruct / pe} & 6.8 & 17.3 & 0.9 & 1.4 & 1.8 & 1.2  \\
    \hline
        \texttt{salamandraTA-7b-instruct / direct} & \cellcolor{blue!10} 39.0 & 15.2 & 13.8 & 3.7 & 41.4 & 26.7  \\
        \texttt{salamandraTA-7b-instruct / icl} & 36.4 & \cellcolor{blue!10} 18.7 & \cellcolor{blue!10} 21.5 & 6.5 & \cellcolor{blue!10} 47.5 & \cellcolor{blue!10} 32.2  \\
        \texttt{salamandraTA-7b-instruct / cot} & 34.4 & 11.0 & 12.7 & \cellcolor{blue!10} 7.3 & 42.5 & 30.0  \\
        \texttt{salamandraTA-7b-instruct / pe} & 4.2 & 7.7 & 0.2 & 0.2 & 0.5 & 0.5  \\
    \end{tabular}
    \label{tab:results_explicit}
\end{table}

	\begin{figure}[H]\centering
		\includegraphics[width=\textwidth]{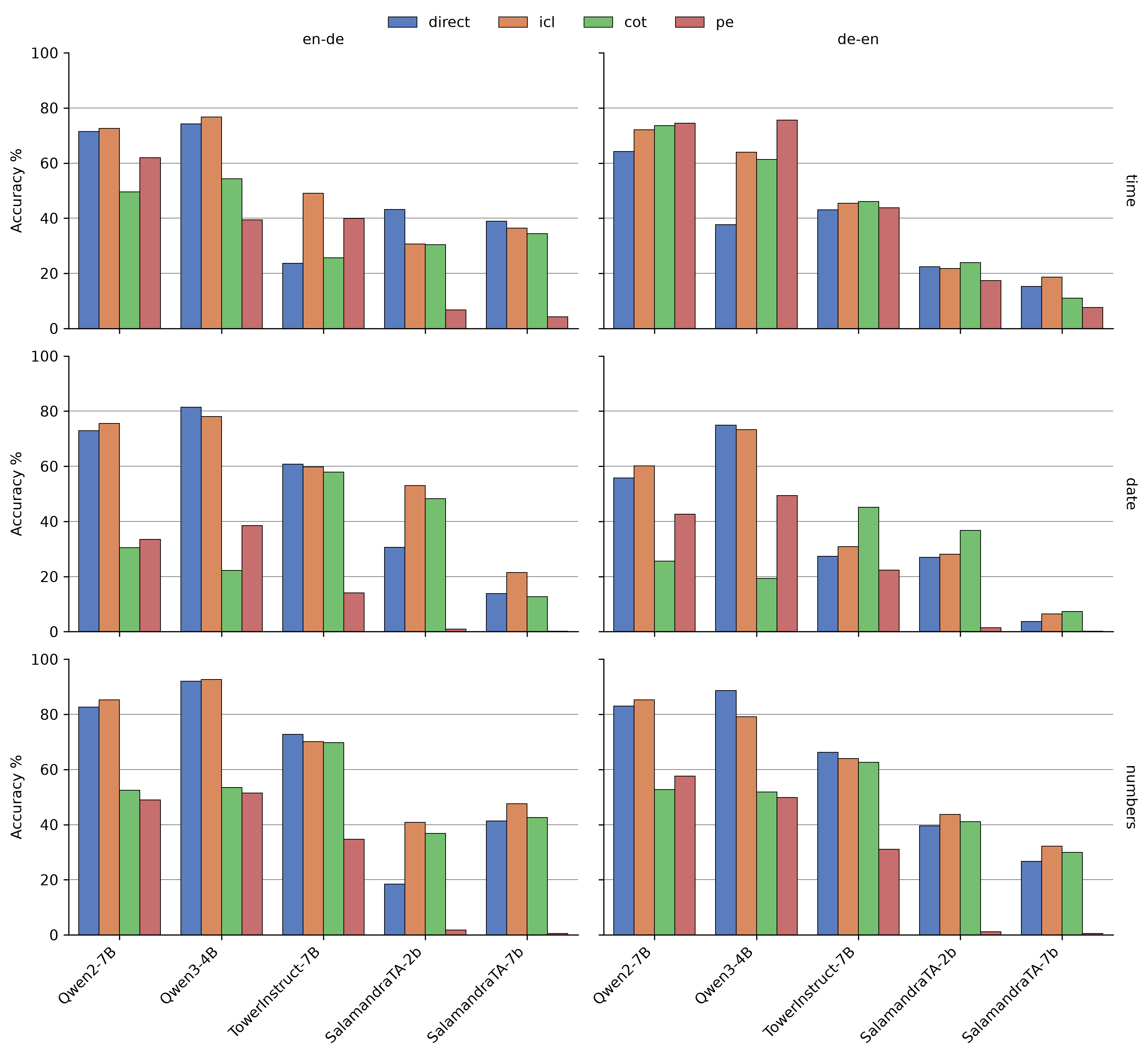}
		\caption{Accuracy for each model and strategy with explicit locale.}
		\label{fig:results_explicit}
	\end{figure}

\subsection{Failure Modes}
	A large number of failures were introduced by spurious adjustment of times, for example, translating \textit{14 Uhr} to \textit{2:30 pm} rather than \textit{2:00 pm}, or the translation only containing the month and year of a date, dropping the day.
	Other failure modes included translating numerals into text, for instance \textit{14.02.2023} as \textit{fourteenth of February twenty twenty three}, which, whilst technically correct, would be unlikely to be deemed acceptable in any but the most flexible of circumstances.

	As shown in table~\ref{tab:model_strategy_accuracies}, the SalamandraTA models performed uniquely poorly with post-editing, and consistently produced responses such as ``We'll release results on Tuesday 23th July @ 2:30pm.'', despite this not being related to the given prompt.
	This is likely to have dragged down the average post-editing accuracy.

	\begin{table}[H]\centering
    \footnotesize
    \caption{Per-model mean accuracy of each strategy across all data types.}
    \begin{tabular}{c | c c | c c | c c | c c}
            & \multicolumn{2}{c|}{Direct}  & \multicolumn{2}{c|}{ICL}  & \multicolumn{2}{c|}{CoT}  & \multicolumn{2}{c}{PE} \\
            & unsp. & expl. & unsp. & expl. & unsp. & expl. & unsp. & expl. \\
        \hline
        \texttt{Qwen2-7B-Instruct}  & 70.91\% & 71.67\% & 74.29\% & 75.15\% & 47.64\% & 47.43\% & 49.60\% & 53.19\% \\
        \texttt{Qwen3-4B-Instruct-2507}  & 78.86\% & 74.82\% & 78.90\% & 77.29\% & 42.46\% & 43.73\% & 52.58\% & 50.71\% \\
        \texttt{TowerInstruct-7B-v0.2}  & 48.71\% & 48.98\% & 55.51\% & 53.20\% & 52.50\% & 51.19\% & 25.90\% & 30.98\% \\
        \texttt{salamandraTA-2b-instruct}  & 29.01\% & 30.16\% & 31.07\% & 36.34\% & 34.10\% & 36.21\% & 4.35\% & 4.89\% \\
        \texttt{salamandraTA-7b-instruct}  & 23.95\% & 23.29\% & 26.91\% & 27.11\% & 24.08\% & 22.98\% & 2.26\% & 2.19\% \\
    \end{tabular}
    \label{tab:model_strategy_accuracies}
\end{table}

	\section{Discussion}

Interestingly, \citeauthor{Tang2025}'s finding that post-editing for numerical translation was the most reliable strategy did not hold for numerical \textit{localisation}.
This is likely due to the nature of localisation as a task of applying formatting rules: when the rules are included in the prompt itself, it makes sense that the results would be improved compared to relying on the implicit knowledge of localisation rules contained within the model.

Our results are consistent with the \citeauthor{Tang2025}'s finding that parameter count is not necessarily indicative of improved performance, as can be seen by the accuracy of \texttt{SalamandraTA-7b} compared to \texttt{SalamandraTA-2b}, shown in Figures~\ref{fig:results_unspecified} and \ref{fig:results_explicit}.

The Salamandra models performed very poorly in comparison to the others: these models had a tendency to change the format of the results, for instance, changing "17/02/2014" to ``seventeenth February two thousand twenty four".
Despite having the correct value, it does not align with the requirements of the task: such examples were not parsed as having the correct value or format.

Explicitly specifying the locale had a small but significant impact on the results: 66/120 cases showed a mean improvement in accuracy of 4.2\%, with 52 showing a mean loss of 3.2\%, with two cases showing no change.
It is intuitive that an explicit locale specification would improve the results, so the limited impact was unexpected.

Some prompt tweaking was helpful in order to account for ancillary text generated by the models.
Attempting to conclusively resolve this problem is an intractable problem: given the statistical nature of LLM-generated text, it is impossible to be certain that all cases are handled.

\subsection{Future Work}
	There are several potential directions for future work. Prompt engineering could be performed for each model to determine the optimal prompt for accuracy, but a systematic study of prompt formats was out of scope for this paper.

	It could also prove interesting to repeat this analysis to determine the optimal size or quantisation within a model family for achieving a given localisation accuracy.

	\pagebreak
	\appendix
	\setcounter{section}{0}
	\section{Appendix}

\subsection{Regular Expressions}
\label{app:regexes}
\begin{footnotesize}
\begin{verbatim}
# Ignore dates without a day as is then just a question of translating the month name, which isn't interesting.
DE_DATE_RE = re.compile(
    r"(?<=\D)[0-9]{4}-[0-9]{1,2}-[0-9]{1,2}(?=\D)|"     # YYYY-MM-DD
    r"(?<=\D)[0-9]{1,2}\.[0-9]{1,2}\.[0-9]{4}(?=\D)|"   # DD.MM.YYYY

    "("    # DD. Monat | DD. Monat YYYY
        r"(?<=\D)[0-9]{1,2}\.? "
         "(Jan(uar)?|Feb(ruar)?|März?|Apr(il)?|Mai|Juni?|Juli?|Aug(ust)?|Sep(tember)?"
         "|Okt(ober)?|Nov(ember)?|Dez(ember)?)"
        r"( [0-9]{4}(?=\D))?"
    ")|"
    "("    # Monat 1234
        "(Jan(uar)?|Feb(ruar)?|März?|Apr(il)?|Mai|Juni?|Juli?|Aug(ust)?|Sep(tember)?"
        "|Okt(ober)?|Nov(ember)?|Dez(ember)?)"
        r"( [0-9]{4}(?=\D))"
    ")"
)

EN_DATE_RE = re.compile(
    r"(?<=\D)[0-9]{4}-[0-9]{1,2}-[0-9]{1,2}(?=\D)|"   # YYYY-MM-DD
    r"(?<=\D)[0-9]{1,2}/[0-9]{1,2}/[0-9]{4}(?=\D)|"   # DD/MM/YYYY

    "("  # DDth Month YYYY | DD Month YYYY
        r"(?<=\D)[0-9]{1,2}(st|nd|rd|th)? (of )?"
         "(Jan(uary)?|Feb(ruary)?|Mar(ch)?|Apr(il)?|May|Jun(e)?|Jul(y)?|Aug(ust)?|Sep(tember)?"
         "|Oct(ober)?|Nov(ember)?|Dec(ember)?)"
        r"(,? [0-9]{1,4}(?=\D))?"
    ")|"
    "("  # Month DDth, YYYY
         "(Jan(uary)?|Feb(ruary)?|Mar(ch)?|Apr(il)?|May|Jun(e)?|Jul(y)?|Aug(ust)?|Sep(tember)?"
         "|Oct(ober)?|Nov(ember)?|Dec(ember)?)"
        r" [0-9]{1,2}(st|nd|rd|th)"
        r"(,? [0-9]{1,4}(?=\D))?"
    ")|"
    "("    # Month 1234
         "(Jan(uary)?|Feb(ruary)?|Mar(ch)?|Apr(il)?|May|Jun(e)?|Jul(y)?|Aug(ust)?|Sep(tember)?"
         "|Oct(ober)?|Nov(ember)?|Dec(ember)?)"
        r"(,? [0-9]{1,4}(?=\D))"
    ")"
)

DE_TIME_RE = re.compile(
    r"(?<=\D)[012]?[0-9] Uhr( [0-5][0-9] Minuten)?|"
    r"(?<=\D)[012]?[0-9]:[0-5][0-9] Uhr( [0-5][0-9] Minuten)?|"
    r"(?<=\D)[012]?[0-9]\.[0-5][0-9] Uhr( [0-5][0-9] Minuten)?"
)

# dateparser bug: https://github.com/scrapinghub/dateparser/issues/1159
DATEPARSER_DE_TIME_BUG_RE = re.compile(r"(?<=\D)[012]?[0-9] Uhr")

EN_TIME_RE = re.compile(
    r"(?<=\D)[01]?[0-9] ?(AM|A\.M\.|am|a\.m\.|PM|P\.M\.|pm|p\.m\.)|"                       # HH am/pm
    r"(?<=\D)[012]?[0-9][:.][0-5][0-9] ?(AM|A\.M\.|am|a\.m\.|PM|P\.M\.|pm|p\.m\.|hours)|"  # HH:MM am/pm
    r"(?<=\D)[012][0-9][:.][0-5][0-9](?=\D)"                                               # HH:MM (24 hour)
    r"(?<=\D)[012][0-9] (o'clock|oclock)"                                                  # 12 o'clock
)

NUMBERS_RE = re.compile(r"[0123456789]+([,.'][0123456789]+)(?! Uhr)")

YEAR_RE = re.compile(r"(?<=\D)19[0-9]{2}(?=\D)|(?<=\D)20[0-9]{2}(?=\D)")
\end{verbatim}
\end{footnotesize}

\section{LLM Prompts}
\label{app:llm_prompts}

\noindent The prompts follow the ChatML format and use the following variables:
\begin{enumerate}
    \item \texttt{SRC\_LANG}: "English" / "German"
    \item \texttt{TARGET}: "English"/"German" or "English (en\_GB)"/"German (de\_DE)", if the locale is specified.
    \item \texttt{SRC\_TEXT}: the sentence in \texttt{SRC\_LANG}.
    \item \texttt{ITEM\_LIST}: a newline-separated list of values.
\end{enumerate}

\subsection{Direct Translation}
\begin{footnotesize}
\begin{verbatim}
{ "role": "system",
  "content": You are an excellent translator.\n"
    "Output only the requested translation with no explanation or additional text"
    " before or after the translation."
},
{ "role": "user",
  "content": f"Translate the following {SRC_LANG} sentence into {TARGET}: {SRC_TEXT!r}."
}
\end{verbatim}
\end{footnotesize}

\subsection{In-context Learning}
\begin{footnotesize}
\begin{verbatim}
{ "role": "system",
  "content": You are an excellent translator.\n"
    "Use the following localisation principles:\n"
    "* German text uses a a comma (,) as its decimal marker and period (.) as a digit separator.\n"
    "* German text uses a period (.) to separate values in purely numerical dates.\n"
    "* German text uses the 24-hour clock.\n"
    "* English text uses a a period (.) as its decimal marker and comma (,) or apostrophe (') as"
        " a digit separator.\n"
    "* English text uses a forward slash, or solidus, (/) to separate values in purely numerical dates.\n"
    "* English text uses either the 24-hour clock, or the 12-hour clock with an AM/PM suffix to"
        " indicate morning or afternoon.\n"
    "Output only the requested translation with no explanation or additional text before or after"
        " the translation."
},
{ "role": "user",
  "content": f"Translate the following {SRC_LANG} sentence into {TARGET} using the"
    f" given localisation principles. {SRC_LANG} sentence: {SRC_TEXT!r}."
}
\end{verbatim}
\end{footnotesize}

\subsection{Chain of Thought}
\begin{footnotesize}
\begin{verbatim}
{ "role": "system",
  "content": "You are an excellent translator.\n"
   f"Translate the {SRC_LANG} sentence into {TARGET} step-by-step.Pay attention to the"
    " localisation conversion between {SRC_LANG} and {TARGET}."
   f"First translate the numerical parts, and then translate the rest of the sentence. "
    "Precede the final translated sentence with '### FINAL TRANSLATION: '"
},
{ "role": "user",
  "content": f"Translate the following {SRC_LANG} sentence into {TARGET}: {SRC_TEXT!r}."
}
\end{verbatim}
\end{footnotesize}

\subsection{Post-Editing}
\texttt{TGT\_TEXT}: the sentence in \texttt{TARGET}, from direct translation.
\begin{footnotesize}
\begin{verbatim}
{ "role": "system",
  "content":
     "You are an excellent extractor of date/time/number translation pairs.\n"
    f"Extract all date/time/number pairs from the given {SRC_LANG}-{TARGET} sentence pairs.\n"
     "Output the extracted translation pairs in the form of list without giving any explanation."
     "Separate items within a pair by '____'.\n"
     "Separate one item in the list from the next by using '####'."
     "Here is an example:\n"
    f"{SRC_LANG}: We will share the result on Tuesday 23rd July at 2:30 pm.\n"
    f"{TARGET}: Wir werden die Resultat am Dienstag 23. Juli freigeben um 14:30 Uhr.\n"
     "output: Tuesday 23rd July ____ Dienstag 23. Juli #### 2:30 pm ____ 14:30 Uhr"
},
{ "role": "user",
  "content":
    f"Here is the {SRC_LANG}-{TARGET} sentence pair from which you need to extract"
     " data/time/number pairs: {SRC_TEXT!r}, {TGT_TEXT!r}"
}
\end{verbatim}
\end{footnotesize}

	\vspace{6pt}

	\funding{This research received no external funding.}
	\dataavailability{The test scripts and data are available at \url{https://doi.org/10.5281/zenodo.21797172}.}
	\acknowledgments{No generative AI was used in the preparation of this manuscript outside the experiments described in section 3.}
	\conflictsofinterest{The author declares no conflicts of interest.}

	\reftitle{References}
	\bibliography{references}
\end{document}